\documentclass{article}

\AtBeginDocument{%
  \providecommand\BibTeX{{%
    \normalfont B\kern-0.5em{\scshape i\kern-0.25em b}\kern-0.8em\TeX}}}

\usepackage[accepted]{mlsys2025}

\usepackage{hyperref}
\usepackage{tikz}
\usepackage{algorithm}
\usepackage{algorithmic}
\usepackage{graphicx}
\usepackage{textcomp}
\usepackage{multirow}
\usepackage{thumbpdf}
\usepackage{booktabs}
\usepackage{colortbl}
\usepackage{url}
\usepackage{subfigure}
\usepackage{array}
\usepackage[normalem]{ulem}
\usepackage{xspace}
\usepackage{wrapfig}
\usepackage{tcolorbox}
\usepackage[framemethod=tikz]{mdframed}
\usepackage{enumitem}
\usepackage{balance}
\usepackage{fontawesome}
\usepackage{pifont}
\usepackage{listings}
\usepackage{fancyvrb}
\usepackage[scaled=0.85]{beramono}
\usepackage{amsmath}
\usepackage{amssymb}
\usepackage{cleveref}
\usepackage{amsthm}
\usepackage[T1]{fontenc}
\usepackage{makecell}
\usepackage{pdflscape}
\usepackage{longtable}
\usepackage{rotating}
\usepackage[nointegrals]{wasysym}

\usepackage{microtype}
\usepackage{graphicx}
\usepackage{booktabs} %

\usepackage{hyperref}

\theoremstyle{definition}
\newtheorem{example}{Example}[section]

\usepackage{longtable} %

\usepackage{titlesec}

\titlespacing{\section}{0pt}{1.5pt plus 0.5pt minus 2pt}{1.5pt plus 0.5pt minus 1pt}
\titlespacing{\subsection}{0pt}{0.5pt plus 0.5pt minus 2pt}{0.5pt plus 0.5pt minus 1pt}
\titlespacing{\subsubsection}{0pt}{0.5pt plus 0.5pt minus 2pt}{0.5pt plus 0.5pt minus 1pt}

\newcommand{\ourbench}[0]{\textsc{AIOpsLab}}

\definecolor{edgeblue}{RGB}{0, 0, 200}
\definecolor{edgegreen}{RGB}{0, 200, 0}
\definecolor{gptgreen}{RGB}{0, 166, 126}
\definecolor{scholarpurple}{RGB}{169, 1, 251}
\definecolor{bgcode}{rgb}{0.95,0.95,0.95}
\definecolor{githubgreen}{rgb}{0.564, 0.933, 0.564}
\definecolor{orange}{rgb}{1,0.5,0}

\definecolor{codegreen}{rgb}{0,0.6,0}
\definecolor{codegray}{rgb}{0.5,0.5,0.5}
\definecolor{backcolour}{RGB}{245,248,250}
\definecolor{emph}{RGB}{166,88,53}
\definecolor{nightblue}{RGB}{9,49,105}
\definecolor{keywords}{RGB}{207,33,46}
\definecolor{lightpurple}{RGB}{130,81,223}
\definecolor{examplebg}{RGB}{250,243,240}
\definecolor{codemph}{RGB}{150,30,30}

\newcommand{\smalltextsc}[1]{\textsc{\small #1}}

\newcommand{\para}[1]{\smallskip\noindent {\bf #1} }

\definecolor{mscolor}{rgb}{0.1,0.1,0.9}

\definecolor{gscolor}{rgb}{0.1,0.8,0.1}

\definecolor{yscolor}{rgb}{0.7,0.3,0.7}

\definecolor{mhcolor}{rgb}{0.7,0.3,0.7}

\definecolor{jmcolor}{rgb}{0.2,0.9,0.9}

\definecolor{pbcolor}{rgb}{0.5,0.5,0.5}

\definecolor{owncolor}{rgb}{0.6,0.6,0.6}

\definecolor{etacolor}{rgb}{0.7,0.7,0.7}

\definecolor{yfcolor}{rgb}{1,0,0}

\definecolor{codegreen}{rgb}{0,0.45,0}
\definecolor{codegray}{rgb}{0.5,0.5,0.5}
\definecolor{codepurple}{rgb}{0.58,0,0.82}
\definecolor{backcolour}{rgb}{0.95,0.95,0.92}
\definecolor{mauve}{rgb}{0.58,0,0.82}

\definecolor{bg}{gray}{0.95}
\colorlet{shadecolor}{gray!20}
\definecolor{Gray}{gray}{0.8}

\newcommand{\code}[1][]{\lstinline[language=Python,basicstyle=\footnotesize\ttfamily,#1]}

\newcommand{\gptthreefiveturbo}{\smalltextsc{GPT-3.5-turbo}}

\newcommand{\gptfourturbo}{\smalltextsc{GPT-4-turbo}}

\newcommand{\gptwshell}{\smalltextsc{GPT-w-Shell}}
\newcommand{\gptthreeshell}{\smalltextsc{GPT-3.5-w-Shell}}
\newcommand{\gptfourshell}{\smalltextsc{GPT-4-w-Shell}}

\newcommand{\react}{\smalltextsc{ReAct}}
\newcommand{\reactname}[0]{\textsc{ReAct}}

\newcommand{\flash}{\smalltextsc{Flash}}

\newcommand{\mksmc}{\smalltextsc{MKSMC}}
\newcommand{\pdiagnose}{\smalltextsc{PDiagnose}}
\newcommand{\rmlad}{\smalltextsc{RMLAD}}

\newcommand{\getlogs}{\texttt{\small get\_logs}}
\newcommand{\getmetrics}{\texttt{\small get\_metrics}}
\newcommand{\gettraces}{\texttt{\small get\_traces}}

\definecolor{gray}{gray}{0.5}
\definecolor{darkred}{RGB}{139, 0, 0}
\definecolor{LightGray}{gray}{0.85}
\definecolor{VeryLightGray}{gray}{0.90}
\definecolor{ForestGreen}{rgb}{0.13, 0.55, 0.13}
\definecolor{Maroon}{rgb}{0.5, 0.0, 0.0}
\definecolor{darkpastelred}{rgb}{0.76, 0.23, 0.13}

\usetikzlibrary{shapes.geometric,arrows,positioning,fit,calc,shadows.blur,shapes.symbols}
\tikzstyle{largerect} = [rectangle, rounded corners, minimum width=8.5cm, minimum height=1cm, font=\footnotesize, text width=8.5cm, fill=red!30]

\tikzstyle{largerectdouble} = [rectangle, rounded corners, minimum width=8.5cm, minimum height=2cm, font=\footnotesize, text width=8.5cm, fill=red!30]

\tikzstyle{smallrect} = [rectangle, rounded corners, minimum width=2.1, minimum height=0.8cm, text width=2.1cm, font=\scriptsize, draw=black, fill=yellow!30]

\tikzstyle{smallrecthalf} = [rectangle, rounded corners, minimum width=1.6cm, minimum height=0.8cm, text width=0.85cm, font=\footnotesize, draw=black, fill=yellow!30]

\tikzstyle{smallrectdouble} = [rectangle, rounded corners, minimum width=2.7cm, minimum height=0.8cm, text width=1.8cm, font=\footnotesize, draw=black, fill=yellow!30]

\tikzstyle{arrow} = [thick,->,>=stealth]

\newenvironment{packed_itemize}{
\begin{list}{\labelitemi}{\leftmargin=1.0em}
 \setlength{\itemsep}{2.5pt}
 \setlength{\parskip}{0pt}
 \setlength{\parsep}{0pt}
 \setlength{\headsep}{0pt}
 \setlength{\topskip}{0pt}
 \setlength{\topmargin}{0pt}
 \setlength{\topsep}{0pt}
 \setlength{\partopsep}{0pt}
}{\end{list}}

\newcommand{\graycircle}[1]{%
    \begin{tikzpicture}[baseline=(textnode.base)]
        \node[circle, draw=none, fill=gray, text=white, font=\bfseries, minimum width=0.37cm, minimum height=0.37cm, inner sep=0pt] (textnode) {#1};
    \end{tikzpicture}%
}

\def\Snospace~{\S{}}

\newcounter{insightC}

\begin{document}

\twocolumn[
\mlsystitle{AIOpsLab: A Holistic Framework to Evaluate AI Agents for Enabling Autonomous Clouds}

\mlsyssetsymbol{equal}{*}

\begin{mlsysauthorlist}
    \mlsysauthor{Yinfang Chen}{uiuc}
    \mlsysauthor{Manish Shetty}{ucb}
    \mlsysauthor{Gagan Somashekar}{ms}
    \mlsysauthor{Minghua Ma}{ms}
    \mlsysauthor{Yogesh Simmhan}{IISc}
    \mlsysauthor{Jonathan Mace}{ms}
    \mlsysauthor{Chetan Bansal}{ms}
    \mlsysauthor{Rujia Wang}{ms}
    \mlsysauthor{Saravan Rajmohan}{ms}
\end{mlsysauthorlist}
    
\mlsysaffiliation{uiuc}{UIUC, Champaign, USA}
\mlsysaffiliation{ucb}{UC Berkeley, Berkeley, USA}
\mlsysaffiliation{IISc}{IISc, Bengaluru, India}
\mlsysaffiliation{ms}{Microsoft, Redmond, USA}

\mlsyscorrespondingauthor{Minghua Ma}{minghuama@microsoft.com}

\mlsyskeywords{Machine Learning, MLSys}

\vskip 0.3in

\begin{abstract}
    AI for IT Operations (AIOps) aims to automate complex operational tasks, 
    such as fault localization and root cause analysis,
    to reduce human workload and minimize customer impact.
While traditional DevOps tools and AIOps algorithms 
    often focus on addressing isolated operational tasks, 
    recent advances in Large Language Models (LLMs) 
    and AI agents are revolutionizing AIOps 
    by enabling end-to-end and multitask automation.
This paper envisions a future where AI agents autonomously 
    manage operational tasks throughout the entire incident lifecycle, 
    leading to self-healing cloud systems, a paradigm we term \textit{AgentOps}. 
Realizing this vision requires a comprehensive framework 
    to guide the design, development, and evaluation of these agents. 
To this end, we present \ourbench{}, 
    a framework that not only 
    deploys microservice cloud environments, 
    injects faults, generates workloads, 
    and exports telemetry data 
    but also orchestrates these components 
    and provides interfaces for interacting with and evaluating agents.
We discuss the key requirements for such a holistic framework 
    and demonstrate how \ourbench{} 
    can facilitate the evaluation of next-generation AIOps agents. 
Through evaluations of state-of-the-art LLM agents within 
    the benchmark created by \ourbench{}, 
    we provide insights into their 
    capabilities and limitations 
    in handling complex operational tasks in cloud environments.

\end{abstract}

]
\printAffiliationsAndNotice{}

\section{Introduction}
\label{sec:intro}
The rapid evolution of IT applications and services 
    has led enterprises to increasingly depend on hyper-scale, 
    cloud-based systems. 
These systems are often distributed, employing architectures such as microservices and serverless computing, enabling scalability but also adding complexity and introducing new operational challenges. 
In such cloud environments, issues can cascade into large-scale outages. 
For instance, an Amazon outage can result in losses of \$100 million in just one hour~\cite{amazonoutage}.

\begin{figure}[t]
    \centering
    \includegraphics[width=1\linewidth]{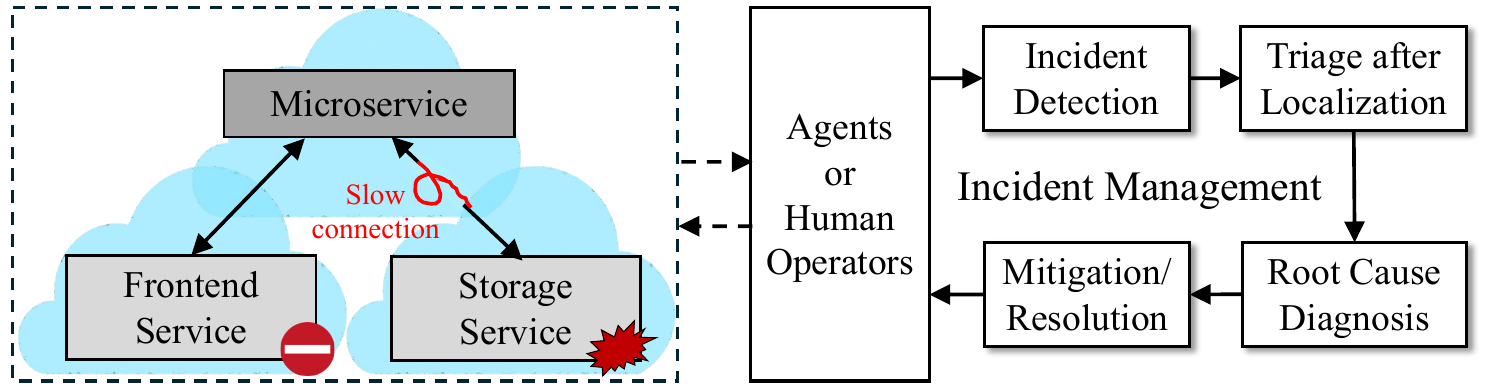}
    \vspace{-7.5pt}
    \caption{Microservice incident and its management lifecycle.} %
    \label{fig:intro}
    \vspace{-7.5pt}
  \end{figure}

To address the challenges of managing incidents 
    in such complex infrastructures, 
    there is a movement towards the adoption of AIOps 
    (Artificial Intelligence for IT Operations),
    within the context of DevOps (Development and Operations). 
The ultimate goal of AIOps is to create autonomous self-healing clouds, 
    where AI-driven approaches can detect, localize, 
    and mitigate faults with minimal human intervention. 
Although such a concept
    has existed for over a decade~\cite{6257956,10.1007/978-3-642-10665-1_5},
    the recent advancements of AIOps and Large Language Model (LLM) agents
    have brought this vision closer to reality~\cite{zhao2023robust, he2022empirical, ma2018robust, Zhang:2018, ganatra2023detection,gamma,zhang2024automated, chen2024automatic}.
Large Language Model (LLM) agents~\cite{mialon2023augmented,schick2024toolformer}
    integrate external tools 
    to dynamically interact with their environment~\cite{wei2022chain},
    enabling them to autonomously manage the entire incident lifecycle, as shown in~\Cref{fig:intro}.

To realize this autonomous self-healing cloud vision, 
    we propose a new paradigm called \textit{AgentOps} (Agent for Operations). 
In this paradigm, 
    agentic approaches are not limited to isolated operational tasks 
    but are capable of seamlessly managing multiple, 
    cross-layer tasks across the entire operational stack. 
AgentOps represents an evolution where autonomous agents 
    can make real-time decisions 
    and end-to-end actions to ensure system reliability. 
This aligns with recent advancements in AI, 
as highlighted by a post:

\begin{quote}
    \vspace{-10.5pt}
    \textit{``State-of-the-art AI results are increasingly obtained 
    by compound systems with multiple components, not just monolithic models ...
    compound AI systems will likely be the best way to maximize AI results in the future''} --
    \textit{\emph{The Shift from Models to Compound AI Systems}}~\cite{compound-ai-blog}
    \vspace{-10.5pt}
\end{quote}

AI-driven tools and benchmarks like WebArena~\cite{webarena}, R2E~\cite{r2e}, HumanEval~\cite{humaneval}, LiveCodeBench~\cite{lcb}, and SWE-bench~\cite{swebench} have significantly advanced the `Dev' side of DevOps by accelerating software development. 
However, progress in AI for `Ops', particularly AgentOps, remains limited, 
    due to the lack of high-quality benchmarks for diverse, realistic scenarios. Addressing this gap requires
    \textit{a framework that aids the design, 
    development, and evaluation of AIOps agents} 
    within an interactive environment, 
    a key contribution of this paper.

\para{Challenges and contributions.}
Building a holistic benchmark framework that 
    can allow agents to interact dynamically with the cloud 
    poses several challenges.
The first challenge is to manage an evaluation flow
    that is generally applicable to diverse agents and clouds, 
    powerful enough to evaluate agents 
    by complex and realistic operational tasks, 
    and valuable enough to provide different feedback or observability, 
    together with extensibility 
    that make it possible to accommodate 
    new tasks and agents by the users.
While existing tools address individual components of the AIOps evaluation, 
    such as observability~\cite{10.1145/3611643.3613881,simonsson2021observability}, 
    application suites~\cite{deathstarbench, trainticket, sriraman2018mu} and 
    chaos engineering~\cite{chaosmonkey, chaosblade, chaosmesh}, 
    they lack the integration necessary 
    to support a unified AIOps evaluation.

We present \textbf{\ourbench{}}, a holistic framework that can automatically manage the entire end-to-end evaluation process for AIOps solutions.
This involves deploying services, fault injection, workload generation, orchestrating the agent-cloud interaction, and analyzing results.
Specifically, \ourbench{} features the \textit{Agent-Cloud Interface (ACI)}, 
    a unified interface that enables agents 
    to interact with the cloud. 
ACI allows agents to communicate, take action, 
    and receive feedback, 
    orchestrating these interactions 
    to detect and resolve issues in dynamic
    and interactive environments.

Moreover, a common challenge in operation benchmarks is 
    the lack of realistic evaluation scenarios, 
    as existing approaches often rely on static datasets,
    such as system metrics~\cite{adbench,anomalybench} that are 
    typically time series data, 
    or on fixed question-answer format~\cite{opseval}. 
Such setups do not capture the dynamic, unpredictable, 
    and evolving nature of real-world cloud environments,
    where workloads and incidents fluctuate over time.
To make matters worse, 
    recent efforts on AgentOps~\cite{wang2023rcagent,zhang2024automated} 
    use proprietary services and datasets. 
Furthermore, existing AIOps approaches and their benchmarks often 
    focus only on isolated aspects of the incident lifecycle, 
    such as anomaly detection \cite{yu2024pre} 
    or fault localization \cite{sun2024art}.
This lacks a cohesive framework to evaluate AIOps agents comprehensively. 
Moreover, it limits support for decision-making 
    that could assist in chaining algorithms 
    or selecting the most suitable agent for a given operation scenario.

To address these limitations, 
    we designed a set of evaluation scenarios, 
    referred to as \textit{problems},
    which replicates realistic incidents 
    within the microservice system. 
\ourbench{}'s problem pool 
    is structured around a task-level taxonomy 
    that categorizes tasks of different problems
    across the incident management lifecycle. 
Our approach ensures that evaluation scenarios 
    go beyond simple performance or crash failures 
    (that cannot be further analyzed or mitigated by the agents), 
    incorporating fine-grained root causes 
    to fully assess the diagnostic 
    and mitigation abilities of AIOps agents. 

\para{Implementation.}
We developed \ourbench{}, an innovative framework for building AgentOps benchmarks to evaluate LLM-based AIOps agents. \ourbench{} utilizes two microservice applications from DeathStarBench~\cite{deathstarbench} as testbeds, along with their workload generators. An extensible fault library, integrated with ChaosMesh~\cite{chaosmesh}, enables diverse fault injections into the system. A telemetry observer, incorporating Prometheus~\cite{prometheus} for metrics, Jaeger~\cite{jaeger} for tracing, and Filebeat~\cite{filebeat} and Logstash~\cite{logstash} for logging,
supports on-disk storage of telemetry data, facilitating evaluations of both traditional AIOps algorithms and agentic solutions. We also integrate Helm and Kubernetes APIs into the \ourbench{}'s orchestrator implementation.

To demonstrate the application of our framework in evaluating LLM-based agents as the benchmark,
we use \ourbench{} to create 48 problems as evaluation scenarios   
    covering different types of AIOps tasks, 
    and register four agents from different types on those problems. 
The agent registration is lightweight, with less than a hundred lines of code to implement.
Our evaluation process reveals distinct challenges agents face across tasks. 

\para{Summary.} This paper makes the following contributions:
\vspace{-1em}
\begin{packed_itemize}
    \item We unravel the requirements and challenges of achieving a holistic framework that supports the design, development, and evaluation of autonomous AIOps agents;
    \item We develop a framework, \ourbench{}, which can not only 
        deploy microservice cloud environments, 
        inject faults, generate workloads, 
        and export telemetry data 
        but also orchestrate these components 
        and provide agent-cloud interfaces 
        for interacting with and evaluating agents.
    \item We leverage the \ourbench{} framework 
        to construct a benchmark suite with 48 problems 
        across different AIOps tasks in an interactive environment
        and evaluate four LLM-based agents.
    \item We provide a detailed analysis of the agents' performance
        and limitations by evaluating them on \ourbench{}.
    \item We will make \ourbench{} publicly available.\footnote{The link will be provided.}
\end{packed_itemize}

\section{\ourbench{}}
\label{sec:architecture}
\begin{figure*}[!t]
    \centering
    \includegraphics[width=1.65\columnwidth]{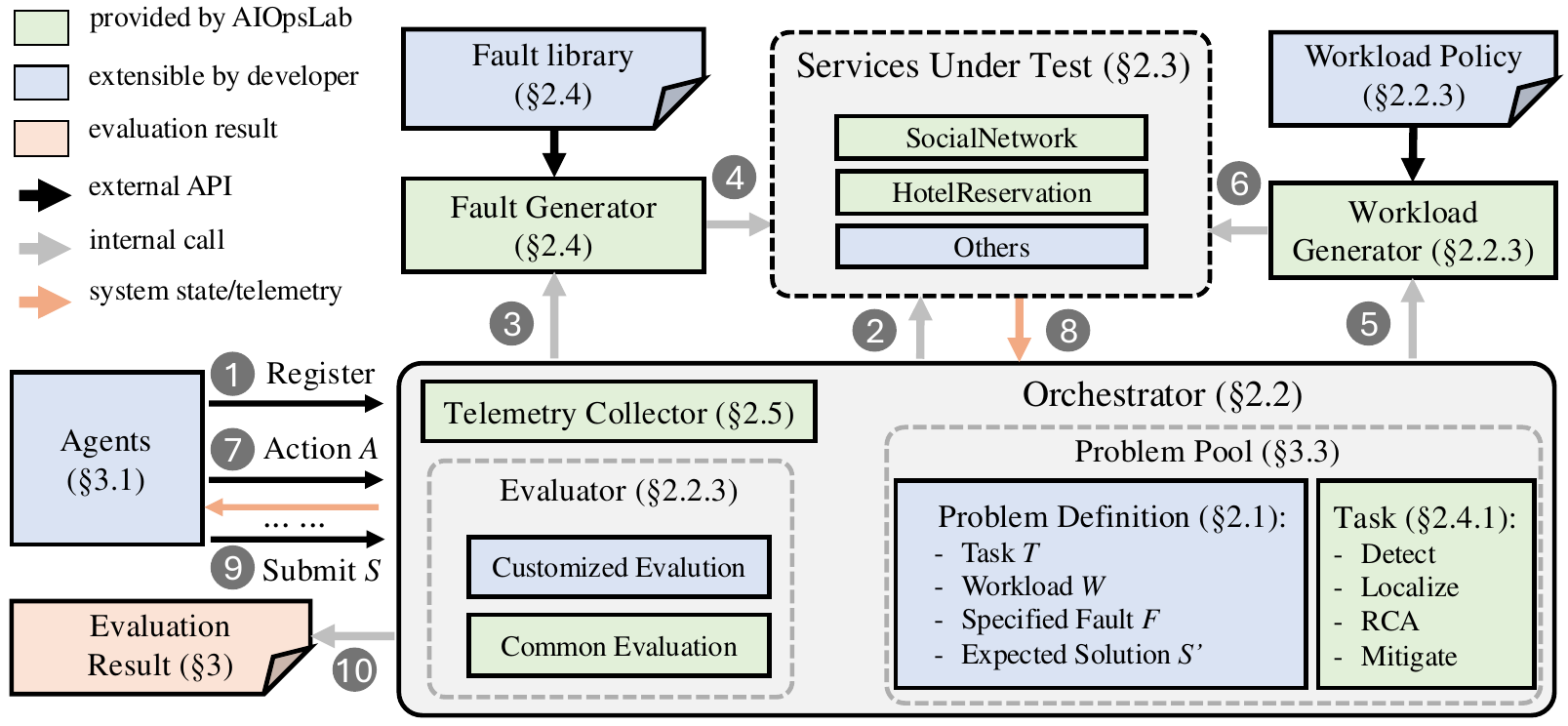}
    \vspace{-7.5pt}
    \caption{\textbf{Overview of \ourbench{}.} \textmd{The Orchestrator coordinates interactions between various system components and serves as the Agent-Cloud-Interface (ACI). 
    Agents engage with the Orchestrator to solve tasks, receiving a problem description, instructions, and relevant APIs. The Orchestrator generates diverse problems using the Workload and Fault Generators, injecting these into applications it can deploy.
    The deployed service has observability, providing telemetry such as metrics, traces, and logs. 
    Agents act via the Orchestrator, which executes them and updates the service's state. The Orchestrator evaluates the final solution using predefined metrics for the task.}
    }
    \label{fig:overview}
    \vspace{-7.5pt}
\end{figure*}

In this section, we discuss the
    design and implementation of \ourbench{} and its components,
    as illustrated in \Cref{fig:overview}.

\subsection{Problem Definition}
\label{subsec:probdef}
To support a wide range of evaluation scenarios (referred to as \textit{problems}),
    which replicate realistic incidents within the microservice system, 
    we first formalize an AIOps problem $P$ as a tuple: $P = \langle T, C, S \rangle$, where
$T$ represents a \textit{task}, $C$ represents a \textit{context}, and $S$ represents the expected \textit{solution} (oracle). 
The task $T$ defines the specific AIOps
operation to be performed, 
    categorized into four types: detection, 
    localization, (root cause) analysis, and mitigation. 
We define these tasks in \Cref{tab:fault_taxonomy}. 
Each task type is associated with success criteria and evaluation metrics. 
For instance, the detection task employs Time-to-Detect (TTD) 
    to measure the time taken to detect a fault.

The \textit{context} $C$ can be further formalized as a tuple: 
    $C = \langle E, I \rangle$, where $E$ is the \textit{operational environment} 
    in which the problem occurs, and $I$ is the \textit{problem information} used to describe the problem to the agent.
The operational environment includes the cloud service, the fault model, and the workload model used to generate the problem, which is not shared with the agent.
The problem information comprises of information 
    such as service descriptions, task descriptions, 
    and documentation about available APIs 
    that is directly shared with the agent.
It also subsumes indirect information (including logs, metrics, and traces observed in the operational environment) that is queryable by the agent at runtime.
Finally, $S$ is the expected outcome of the task, 
    which is used to evaluate the agent's performance. 
The solution is typically problem and task-specific 
    and is carefully designed for evaluation. 
Note that some problems, e.g., mitigation tasks, can be solved in multiple ways.
In such cases, \ourbench{} evaluates the general state of the entire system,
    e.g., check whether all of the services are up and running,
    after the problem is resolved, 
    rather than solely on the targeted resource 
    where the fault was injected, 
    because other services or resources 
    may have been inadvertently affected 
    during the mitigation process.

\begin{example}\label{ex:prob}
    Consider the problem of localizing a Kubernetes target port misconfiguration in a social network application. 
    \ourbench{} makes it easy to define this problem in just a few lines by extending the \code{LocalizationTask} interface.

\begin{lstlisting}
from aiopslab import LocalizationTask, SocialNetwork
from aiopslab import Wrk, VirtFaultInjector
class K8STargetPortMisconf(LocalizationTask):
    def __init__(self):
        self.app = SocialNetwork()
        self.ans = "user-service"

    def start_workload(self):
        wrk = Wrk(rate=100, duration=10)
        wrk.start_workload(url=self.app.frontend_url)
    
    def inject_fault(self):
        inj = VirtFaultInjector(self.app.ns)
        inj.inject([self.ans], "misconfig_k8s")
    
    def eval(self, soln, trace, duration):
        res["TTL"] = duration
        res["success"] = is_exact_match(soln, self.ans) 
        return res
\end{lstlisting}

Here, the task $T$ is fault localization, and the solution $S$ is the microservice named ``user-service'', which is also the fault injection target.
The context $C$ includes the social network application, 
    a misconfiguration fault from \ourbench{}'s fault library,  
    and a standard workload using the \texttt{wrk} tool. 
\ourbench{} provides several such interfaces 
    for all AIOps tasks (\Cref{subsubsec:fault_tax}) 
    and allows users to add new problems by extending them. 
Once problems are defined, 
    \ourbench{} can instantiate them and allow agents to interact 
    with them using an Orchestrator that we describe next.
\end{example}

\subsection{Orchestrator}
\label{subsec:orchestrator}

\ourbench{}'s Orchestrator strictly enforces the separation of concerns 
    between the agent and the service, 
    using a well-defined central piece, 
    the Orchestrator. 
It provides a robust set of interfaces 
    that allow seamless integration 
    and extension of various system components.

\subsubsection{Agent Cloud Interface}
A key responsibility of the Orchestrator is to provide a well-defined interface for the agent to interact with the cloud environment.
Typically, developers operate clouds and services with various programming (e.g., APIs, CLIs) and user interfaces (incident portals, dashboards, etc.).
However, existing interfaces to the cloud are not well-designed for LLMs and agents. For instance, humans can reliably
ignore irrelevant information, which can prove distracting for agents and hamper performance.

The ACI specifies (1) the set of valid actions available to the agent \graycircle{7} \graycircle{9}, 
and (2) how the service's state is conveyed back to the agent as the observation of its actions \graycircle{8}.

In doing so, the ACI abstracts the cloud environment's complexity, simplifying the agent's decision-making process.
The ACI is designed to be intuitive and easy to use, with a concise list of APIs, 
each documented to ensure that agents can make meaningful progress towards their objectives.
Some APIs that \ourbench{} provides by default include \texttt{\small get\_logs} (fetch logs), 
\texttt{\small get\_metrics} (fetch metrics), \texttt{\small get\_traces} (fetch traces),
and \texttt{\small exec\_shell} (execute shell commands after applying security policy filters).

\begin{example}
    This example illustrates how the ACI is defined in \ourbench{} as APIs that agents can use.
\vspace{-1em}
\begin{lstlisting}
class TaskActions:            
    def get_traces(ns: str, duration: int = 5) -> str:
        """
        Collects trace data of the services from Jaeger.
        Args:
            ns (str): The K8S namespace.
            duration (int): Duration to collect traces.
        Returns:
            str: Path to the directory where traces saved.
        """
        trace_api = TraceAPI(ns)
        end_t = datetime.now()
        start_t = end_t - timedelta(duration)
        traces = trace_api.extract_traces(start_t, end_t)
        return trace_api.save_traces(traces)
\end{lstlisting}
\vspace{-1em}
As shown, the ACI encapsulates complex operations 
    behind simple APIs like \texttt{\small get\_traces}.
On initializing a problem, the Orchestrator automatically 
    extracts documentation from these APIs 
    to provide as \textit{context} $C$ 
    to the agent.
At runtime, agents can specify a wide range of actions on the service (e.g., scaling, redeploying, patching) 
by way of the Orchestrator's privileged access.
Finally, the Orchestrator conveys the service's state after each action with high-quality feedback to the agent,
including outputs, error messages, and tracebacks.
\end{example}

\subsubsection{Session Interface}
Another key responsibility of the Orchestrator is to manage the lifecycle of the agent and the service.
We implement the Orchestrator as a session-based system, where a \texttt{\small Session} 
is created for each instance of an agent solving a problem.
Agents are registered with the Orchestrator, and a session starts with simple API calls passing a unique problem identifier \graycircle{1}.
\ourbench{}'s design is highly flexible and integrates with the growing LLM and agent framework space.
Our only requirement is that the agent must implement a \texttt{\small get\_action} method with 
the following signature: \code{async def get_action(state: str) -> str}.
It takes the service's state as input from the Orchestrator and returns the next action the agent wants to take. 
Note that this could be a simple wrapper function around any existing agent framework.

\begin{example}
    In this simplified example, we illustrate how an Agent can be onboarded to \ourbench{}.

\begin{lstlisting}
from aiopslab import Orchestrator
class Agent:
    def __init__(self, prob, instructs, apis):
        self.prompt = self.set_prompt(prob, instructs, apis)
        self.llm = GPT4()

    async def get_action(self, state: str) -> str:
        return self.llm.generate(self.prompt + state)

#initialize the orchestrator
orch = Orchestrator()
pid = "misconfig_app_hotel_res-mitigation-1"
prob_desc, instructs, apis = orch.init_problem(pid)
#register and evaluate the agent
agent = Agent(prob_desc, instructs, apis)
orch.register_agent(agent, name="myAgent")
asyncio.run(orch.start_problem(max_steps=10))
\end{lstlisting}
    \vspace{-1em}
    As shown on initializing a problem, the Orchestrator shares \textit{context} necessary for the agent to solve the problem.
    It then polls (via \texttt{\small get\_action}) for the agent's next action.
\end{example}

\subsubsection{Other Interfaces}
\label{subsub:other_interface}
\textit{Problem Initializers.} 
As described in \Cref{subsec:probdef}, 
    each problem is defined with a \textit{context} $C$
    which includes its operational environment. This environment is the service, fault, and workload conditions under which the problem occurs.
Here, the Orchestrator deploys services and uses infrastructure-as-code tools like \cite{helm} to deploy the required cloud service for each problem. 
We describe services already integrated 
    into \ourbench{} in \Cref{subsec:services}.

As shown in \Cref{fig:overview}, to create realistic benchmark scenarios, the Orchestrator then interfaces with two entities: (1) a \textit{workload generator} \graycircle{5} and (2) a \textit{fault generator} \graycircle{3}.
These generators introduce controlled service disruptions that simulate live benchmark problems.
As the workload generator, \ourbench{} currently uses the \code{wrk2} tool \cite{deathstarbench}, which supports several workload policies and also replays industry workloads \graycircle{6}. However, the \ourbench{} is extensible to other workload generators. For fault generation, \ourbench{} uses a custom fault library that instantiates faults across different levels of the system stack \graycircle{4}, such as application and virtualization. The library contains and extends to several fine-grained and parametric faults that go beyond surface-level symptoms and engage deeper into more complex resolution strategies.
We describe the fault library in detail in \Cref{subsec:fault_lib}.

\textit{Problem Evaluators.} Finally, the Orchestrator plays a critical role in evaluating the agent's performance on a problem. It compares the agent's solutions against predefined success criteria and evaluation metrics specific to each task \graycircle{10}. \ourbench{} supports several default and common metrics for each task (e.g., Time-to-Detect for detection, number of steps taken, and tokens produced by an LLM-powered agent sent to \ourbench{}). 
    Additionally, \ourbench{} provides an optional qualitative evaluation 
    of agent trajectories using LLMs-as-Judges \cite{llmjudge}.
Beyond that, all user-defined evaluation metrics specific to the problem are run. For instance, for the localization problem in \Cref{ex:prob}, the metric \code{success} is defined by the agent's submission matching the fault microservice's name. Lastly, the Orchestrator maintains comprehensive logs of all agent trajectories, including actions taken and resulting system states, 
facilitating detailed analysis and debugging.
All of the evaluation results will be automatically collected.

\subsection{Cloud Services}
\label{subsec:services}
\ourbench{} deploys live microservice applications as cloud environments \graycircle{2}.
\ourbench{} is currently integrated with the HotelReservation and SocialNetwork from DeathStarBench~\cite{deathstarbench}. 
The SocialNetwork application has 28 microservices, including Memcached, MongoDB, and Redis, that together implement several features of real-world social networking applications. 
The HotelReservation application, implemented with Go and gRPC, supports services like recommending and reserving hotels.

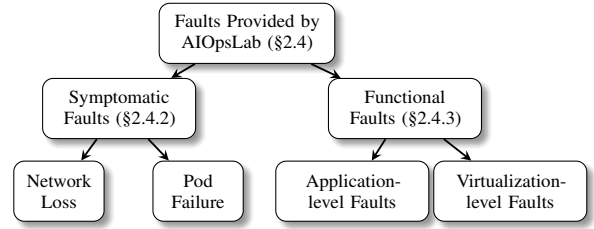
\begin{figure}[!t]
    \centering
    \vspace{-15pt}
    \begin{tikzpicture} [node distance=1.5cm, color=black]
  
      \node (invisible) [rectangle, rounded corners, minimum width=6.5cm, minimum height=2cm, font=\scriptsize, text width=6.5cm] {};
      
      \node (faults) [smallrect, rounded corners, minimum width=1.2cm, minimum height=0.8cm, font=\scriptsize, fill=white, blur shadow={shadow blur steps=5}, align=center] at ([xshift=-1.0cm]invisible.center){Faults Provided by AIOpsLab (\S\ref{subsec:fault_lib})};

      \node (symptomatic) [smallrectdouble, yshift=-1.0cm, rounded corners, minimum width=1.2cm, minimum height=0.8cm, font=\scriptsize, fill=white, blur shadow={shadow blur steps=5}, align=center] at ([xshift=-2.7cm]invisible.center){Symptomatic Faults (\S\ref{subsubsec:symptom})};
      
      \node (functional) [smallrectdouble, yshift=-1.0cm, rounded corners, minimum width=1.2cm, minimum height=0.8cm, font=\scriptsize, fill=white, blur shadow={shadow blur steps=5}, align=center] at ([xshift=1.1cm]invisible.center){Functional Faults (\S\ref{subsubsec:functional})};

      \node (cpu) [smallrecthalf, yshift=-2.1cm, rounded corners, minimum width=1.2cm, minimum height=0.8cm, font=\scriptsize, fill=white, blur shadow={shadow blur steps=5}, align=center]  at ([xshift=-3.5cm]invisible.center) {Network Loss};

      \node (pod) [smallrecthalf, yshift=-2.1cm, rounded corners, minimum width=1.2cm, minimum height=0.8cm, font=\scriptsize, fill=white, blur shadow={shadow blur steps=5}, align=center]  at ([xshift=-1.65cm]invisible.center) {Pod Failure};

      \node (application) [smallrectdouble, yshift=-2.1cm, rounded corners, minimum width=1.2cm, minimum height=0.8cm, font=\scriptsize, fill=white, blur shadow={shadow blur steps=5}, align=center]  at ([xshift=0.40cm]invisible.center) {Application-level Faults};

      \node (virtualization) [smallrectdouble, yshift=-2.1cm, rounded corners, minimum width=0.2cm, minimum height=0.8cm, font=\scriptsize, fill=white, blur shadow={shadow blur steps=5}, align=center]  at ([xshift=2.5cm]invisible.center) {Virtualization-level Faults};
  
      \draw [arrow,line width=0.7pt] (faults) -- (symptomatic);
      \draw [arrow,line width=0.7pt] (faults) -- (functional);
      \draw [arrow,line width=0.7pt] (symptomatic) -- (cpu);
      \draw [arrow,line width=0.7pt] (symptomatic) -- (pod);
      \draw [arrow,line width=0.7pt] (functional) -- (application);
      \draw [arrow,line width=0.7pt] (functional) -- (virtualization);
      
    \end{tikzpicture}
  
    \caption{Fault categories to instantiate problems in \ourbench{}.}
    \label{fig:aiopslab_faults}
    \vspace{-15pt}
\end{figure}

\subsection{Task-oriented Fault Library}
\label{subsec:fault_lib}

\subsubsection{Task Taxonomy}
\label{subsubsec:fault_tax}
We present a task-level taxonomy (\Cref{tab:fault_taxonomy}) that categorizes the tasks that AIOps agents should accomplish according to the different stages of the incident management lifecycle, with progressively increasing complexity.
In \Cref{tab:fault_taxonomy}, a higher level indicates a harder and more impactful task to evaluate agents.

Level 1 focuses on the preliminary identification of unusual behavior within the system, for example, detecting a malfunctioning Kubernetes pod of a microservice.
Also, users can define more complex tasks or create sub-tasks.
The root cause analysis task has both the system level
    and fault type prediction sub-tasks to be solved.

To instantiate problems across different task levels, 
    we use fault injection to inject faults into the system,
    and construct a problem pool for \ourbench{}.  
We classify them into two main types, 
    symptomatic faults and functional faults, 
    as shown in~\Cref{fig:aiopslab_faults}.

\begin{table}[t]
\centering
\caption{Task taxonomy for AIOps agent evaluation. \textmd{The lower the level, the easier the task. \ourbench{} aims to evaluate agents across all task levels with its problems.}}
\label{tab:fault_taxonomy}
\scriptsize
\begin{tabular}{l|p{2cm}|p{4cm}}
\toprule
\textbf{Level} & \textbf{Task (\# sub tasks)} & \textbf{Evaluation Focus}                                                   \\ \hline
1 & Detection (1) & Can the approach accurately detect anomalies or deviations?                                      \\ \hline
2 & Localization (1) & Can the approach pinpoint a fault's exact source (e.g., microservice)? \\ \hline
3 & Root Cause Analysis (RCA) (2) & Can the approach determine the underlying cause of the fault? \\ \hline
4 & Mitigation (1) & Can the approach give effective solutions to recover the environment? \\ \hline
\end{tabular}
\vspace{-5pt}
\end{table}

\begin{figure}[t]
    \centering
    \includegraphics[clip,width=0.65\columnwidth]{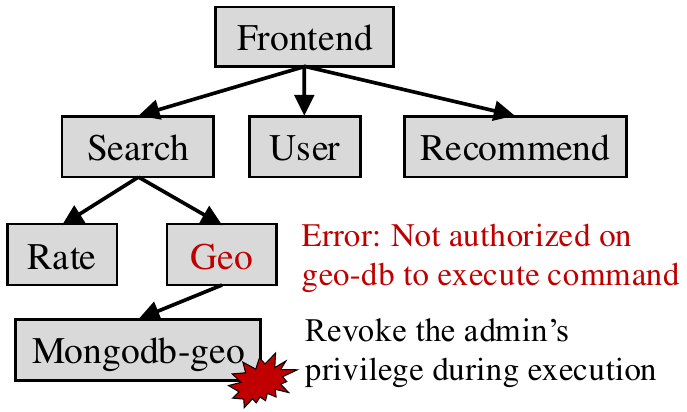}
    \vspace{-15pt}
    \caption{\textmd{Revoke authentication fault example. Injection happens at Mongodb-geo service, while Geo service will be abnormal and generate error logs.}
    }
    \label{fig:revoke-auth}
    \vspace{-15pt}
\end{figure}

\subsubsection{Symptomatic Faults} 
\label{subsubsec:symptom}
Symptomatic faults, such as performance degradation and crash failures, 
    manifest as observable symptoms, 
    such as increased latency, 
    resource exhaustion, or service outages. 
These faults typically help to construct Level 1 
    and Level 2 tasks in the taxonomy, 
    which can create problems that 
    evaluate AIOps approaches' detection and localization ability. 
These faults provide an overview of potential problems 
    but do not necessarily reveal the deeper, 
    underlying root causes of issues (since they do not have one).
\ourbench{} integrates the fault injection tool, 
    Chaos-Mesh~\cite{chaosmesh}, 
    to inject symptomatic faults into microservice applications.

\subsubsection{Functional Faults}
\label{subsubsec:functional}
Though there are many fault injection tools for testing the resilience of cloud systems~\cite{marinescu2009lfi,Banabic2012,Christakis2017a,zhang2012amplifying,jepsen, Pillai2014,alquraan2018analysis,lu2019crashtuner, chen2020cofi, leesatapornwongsa2014samc, gunawi2011fate, majumdar2018why, Ju2013, heorhiadi2016gremlin,alagappan2016correlated,Mohan2018,sun:osdi:22,canini2012nice}, 
    most of them focus solely on injecting system symptoms. 
These coarse-grained faults can only disrupt without modeling the underlying, fine-grained root causes, e.g., misconfigurations or software bugs,
and hence are unable to evaluate the capabilities of AIOps agents 
to diagnose and mitigate root causes.

The failure scenarios to evaluate AIOps agents across tasks must go beyond simple performance or crash failures, and reflect realistic cases that challenge agents, where functional faults come into play.
Functional faults require approaches 
    to not only detect (Level 1) and localize (Level 2) the failure 
    but also diagnose the root cause (Level 3), 
    such as incorrect deployment or operations, 
    and apply the correct mitigation strategies (Level 4). 
For instance, the fault in \Cref{fig:revoke-auth} 
    revokes the admin authentication for the MongoDB database of the geographic microservice (Mongodb-geo). 
Since the Geo service relies on its backend database, 
    errors will appear during its invocation.

\begin{example}
    In the following example, we illustrate the structure of the application-level fault injector for a revoke authentication fault and its usage example in \ourbench{}.
\begin{lstlisting}
from aiopslab.generators.fault.base import FaultInjector
from aiopslab.service.apps.hotelres import HotelReservation
class ApplicationFaultInjector(FaultInjector):
    def inject_revoke_auth(self, microservices: list[str]):
        """Revoke MongoDB admin privileges."""
        ...
    def recover_revoke_auth(self, microservices: list[str]):
        """Recover the revoke admin privileges fault."""
        ...
# Usage Example
class MongoDBRevokeAuth:
    def __init__(self):
        self.app = HotelReservation()
    
    def inject_fault(self):
        injector = ApplicationFaultInjector(ns)
        injector._inject(["mongodb-geo"], "revoke_auth")
\end{lstlisting}
Users can define problems using the existing fault library. For instance, users can specify different faulty services or even construct a task that injects multiple faults into multiple services concurrently. Users can also customize their faults to generate various problems.
\ourbench{} provides the injection function 
    for its associated failure scenarios 
    and offers the corresponding mitigation mechanism 
    to recover the system from the erroneous state. 
In \Cref{subsec:problem_cases}, we will discuss the current problem pool we implement.
\end{example}
\vspace{-0.5em}

\subsection{Observability}
\ourbench{} is equipped with an extensible observability layer to provide comprehensive monitoring capabilities.
\ourbench{} collects a wide array of telemetry data by its telemetry collector, 
    including (1) traces from Jaeger~\cite{jaeger} detailing
the end-to-end paths of requests through distributed systems, (2)
application logs retrieved by Kubectl, or formatted and recorded by Filebeat~\cite{filebeat} and Logstash~\cite{logstash},
and (3) system metrics monitored by Prometheus~\cite{prometheus}.
\ourbench{} not only supports data collection during the interaction with the LLM agent but can also export the data offline to facilitate evaluating other traditional AIOps approaches.
Besides, \ourbench{} is designed to capture information from other dimensions, e.g., codebase, configuration, and cluster information. Developers can also design and expose low-level system information (such as syscall logs) to agents using \ourbench's interface.

\section{Evaluation}
\label{sec:results}
\begin{table*}[htbp]
\caption{\textbf{Selected faults used to instantiate the problems for evaluation in \ourbench{}.} \textmd{Ext. stands for extensibility. $\CIRCLE$ denotes the fault can be easily used to construct other problems; $\LEFTcircle$ denotes there is some manual effort needed to create new problems; while $\Circle$ means the fault is specific to some problems and cannot be applied to create other problems.}}
\vspace{2pt}
\scriptsize
\centering
\resizebox{0.9\linewidth}{!}{
  \begin{tabular}{cl|lccccl}
  \toprule
  \textbf{No.} &
    \textbf{Name} &
    \textbf{Application} &
    \textbf{Task Level} &
    \textbf{Category} &
    \textbf{Ext.} &
    \textbf{\# Problem} &
    \textbf{Description} \\ \midrule

  1 & AuthenticationMissing & HotelReservation & 1, 2, 3, 4 & \makecell[c]{Functional \\ Virtualization} & $\LEFTcircle$ & 4 &
  \makecell[l]{Missing authentication credentials cause \\ access denial to MongoDB.} \\ \hline
  
  2 & TargetPortMisconfig & SocialNetwork & 1, 2, 3, 4 & \makecell[c]{Functional \\ Virtualization} & $\CIRCLE$ & 12 &
  \makecell[l]{The service cannot connect to the specified \\ port due to misconfiguration.} \\ \hline

  3 & RevokeAuth & HotelReservation & 1, 2, 3, 4 & \makecell[c]{Functional \\ Application} & $\LEFTcircle$ & 8 &
  \makecell[l]{Revoked authentication causes database \\ connection failure.} \\ \hline

  4 & UserUnregistered & HotelReservation & 1, 2, 3, 4 & \makecell[c]{Functional \\ Application} & $\LEFTcircle$ & 8 &
  \makecell[l]{The database service has access failures \\ after the user was unregistered.} \\ \hline

  5 & BuggyAppImage & HotelReservation & 1, 2, 3, 4 & \makecell[c]{Functional \\ Application} & $\Circle$ & 4 &
  \makecell[l]{Connection code bug in the application \\ image causes access issues.} \\ \hline

  6 & ScalePod & SocialNetwork & 1, 2, 3, 4 & \makecell[c]{Functional \\ Virtualization} & $\CIRCLE$ & 4 &
  \makecell[l]{Incorrect scaling operation makes the \\ number of pod zero for a service.} \\ \hline

  7 & AssignNonExistentNode & SocialNetwork & 1, 2, 3, 4 & \makecell[c]{Functional \\ Virtualization} & $\CIRCLE$ & 4 &
  \makecell[l]{Pod in a pending a failure status due to \\ wrong assignment to a non-existent node.} \\ \hline
 
  8 & NetworkLoss & \makecell[l]{HotelReservation} & 1, 2 & Symptomatic & $\CIRCLE$ & 2 &
  \makecell[l]{Network loss causes communication \\ failures for a specific service.} \\ \hline

  9 & PodFailure & \makecell[l]{HotelReservation} & 1, 2 & Symptomatic & $\CIRCLE$ & 2 &
  \makecell[l]{Service interruption due to a pod failure.} \\ \hline

  10 & Noop & \makecell[l]{HotelReservation \\ SocialNetwork} & 1 & - & $\CIRCLE$ & 2 &
  No faults injected into the system. \\ \bottomrule
  \end{tabular}
  \label{tbl:faults}
}
\end{table*}

This section begins by outlining the evaluation setup 
    and metrics employed within \ourbench{}. 
We then delve into the selected faults listed in \Cref{tbl:faults}, 
    which serve as diverse evaluation scenarios within \ourbench{}. 
Following this, we evaluate the performance of the AIOps agents 
    solving these problems,
    and then analyze the cost of the agents.
We also dig into the reasons behind the performance differences 
    to understand the challenges and potential agent improvements.
Note that, all of the results are automatically 
    collected and recorded by the problem evaluators (\Cref{subsub:other_interface}).

\subsection{Evaluation Setup}
\label{sec:setting}

We evaluate four LLM-based agents with \ourbench{}. 
Note that, for a fair comparison, 
    we register the naive agent in \ourbench{} 
    without any fine-tuning or modifications.  
We use \gptthreefiveturbo{} and \gptfourturbo{}~\cite{gpt4} that have access to 
    only a secure shell as baselines (\gptwshell{}).
In addition, we also evaluate the performance of \react{}~\cite{react}, 
    which extends chain-of-thought reasoning \cite{cot} 
    by integrating reasoning and acting in an interleaved manner,


As for cloud operation-specific agents, 
    we choose 
    \flash{}~\cite{zhang2024flash}.
\flash{} employs a workflow automation system 
    that monitors execution status and decomposes complex instructions 
    into manageable, conditional segments. 
It incorporates hindsight generation to 
    learn from past interactions. 
As \flash{} was not publicly available at the time of writing, 
    we develop a simplified version that 
    retrospectively generates insights after each step.

To compare with other AIOps approaches specific to a certain type of task, 
    we evaluate three state-of-the-art, 
    non-LLM-based AIOps algorithms on \ourbench{}, 
    using (multi-modal) telemetry data as input.
They are: MKSMC~\cite{ccetin2020anomaly} for detection, 
    RMLAD~\cite{wang2020root} and PDiagnose~\cite{hou2021diagnosing} 
    for localization.

\subsection{Metrics}
\textit{Correctness.}
This metric measures the accuracy of the agent's response to problems. It evaluates whether the agent successfully detects, localizes, analyzes and resolves the problems as expected. 

\textit{Time/Steps.}
These metrics evaluate the efficiency of the AIOps agent for each type of task. For example, Time-to-Detect (TTD) is the time elapsed from the occurrence of a fault to its detection, and Time-to-Mitigate (TTM) is the time taken from detection to complete mitigation of the fault. 
The number of steps or actions taken to solve the problem is also recorded. 
Note that this is the number of times the agent interacts with the \ourbench{} instead of the number of requests sent to the backend LLM.

\textit{Cost.} We use the number of tokens, 
    including both the input token and output tokens, 
    generated by the agents/environment 
    as an indicator of the cost.

\subsection{Problem Pool of \ourbench{} Benchmark}
\label{subsec:problem_cases}

Currently, \ourbench{} benchmark consists of 48 problems in its problem pool. 
With six agents, we evaluate a total of 288 cases.
\Cref{tbl:faults} lists the faults used to instantiate the problems.
As shown in \Cref{tbl:faults}, all functional faults (including Fault 1-7) 
    are used to create problems at all of the four task levels;
    while the symptomatic faults (including Fault 8-9)
    can only be used to create problems at the detection 
    and localization levels (Level 1 and Level 2).
In the detection-level task, the agents must identify the presence of faults in real-time.
This task is a binary classification, where the agents have to respond either ``yes'' if a fault is present or ``no'' on the contrary.
The detection task (Level 1) can be made more complex, e.g., by asking the agents to label the abnormal telemetry data; however, we keep it simple here and leave the complex tasks to other levels. 
The localization (Level 2) task asks the agents to 
    specify the exact location of the fault, 
    usually a service or pod name in Kubernetes.
The RCA task (Level 3) requires the agents to identify (1) the system layer the fault affects and (2) the type of the fault, e.g., misconfiguration or operation error.
The mitigation task (Level 4) requires the agents to 
    \textit{interact with the environment} 
    to fix the fault with a series of actions, 
    such as updating the configuration, 
    or rollback to a previous version, etc. 

Most faults enable users to extend and create new problems easily
    by injecting the fault into other targets, such as services.
For example, Fault 2 in \ourbench{} can be injected into 10 services
    by simply configuring the injection target. 
We select the ``user-service'', ``text-service'', 
    and ``post-storage-service'' from SocialNetwork as injection targets.
Injecting faults into different targets is crucial 
    because each service may have distinct dependencies, 
    resulting in varied fault ``blast radius'' or failure propagation topologies.
Consequently, faults can manifest at different locations 
    within the microservice architecture 
    to help evaluate the ability of the AIOps agents 
    since different locations may indicate distinct difficulties.
Applying some faults to construct problems 
    may require additional effort.
For example, Fault 3 and Fault 4 require the users to not only
    prepare the scripts to trigger the admin privilege 
    revoke or user unregisteration during the testing,
    but also update the config map of the application in Kubernetes;
    and Fault 1 needs to enforce its TLS requirements 
    through a Helm configuration update.
Furthermore, some faults are designed for specific problems 
    and are not readily adaptable, such as Fault 5, 
    which involves an application-level code bug 
    in the microservice's image.

\subsection{Performance Results}
\label{subsec:performance}
The overall performance of the agents is summarized in
    \Cref{table:agent_comparison_overview}, 
    with task-specific results in \Cref{table:agent_performance}.
As illustrated in \Cref{table:agent_comparison_overview}, 
    \flash{} achieves the highest accuracy among all agents. 
Although \gptthreefiveturbo{} completes the tasks the fastest,
    it has the lowest accuracy at 15.25\%.

\begin{table}[t!]
    \centering
    \vspace{-10pt}
    \caption{\textbf{Overall performance of different agents.} \textmd{We show the lines of code (LoC) to register the agent in \ourbench{}, average running time in seconds, average number of steps taken, average tokens used, and accuracy across all problems.}
    }
    \vspace{4pt}
    \scriptsize
    \resizebox{\columnwidth}{!}{
    \begin{tabular}{lrrrrr}
    \toprule
    \textbf{Agent}        & \textbf{LoC} & \textbf{Time (s)} & \textbf{\# Steps} & \textbf{Tokens} & \textbf{Acc.} \\ \midrule
    \gptfourshell{}           & 41 & 28.61 & 6.44  & 6,394.5 & 49.15\%    \\ 
    \gptthreeshell{}          & 41 & 12.44 & 14.70 & 2,557.95 & 15.25\% \\ 
    \react{}                  & 49 & 43.79 & 11.50 & 16,941.46 & 55.93\% \\
    \flash{}                  & 60 & 99.64 & 8.48  & 6,484.25 & \cellcolor{Gray}59.32\%  \\ 
    \bottomrule
    \end{tabular}
    }
    \label{table:agent_comparison_overview}
    \vspace{-10pt}
\end{table}

\begin{table*}[htbp]
    \centering
    \caption{\textbf{Agent performance by task.} \textmd{This table summarizes the performance of different agents across various tasks including detection, localization, RCA, and mitigation. Acc. stands for accuracy. 
    Input/Output represents the number of tokens given to and produced by the agent, respectively.
    }
    }
    \label{table:agent_performance}
    \vspace{-8pt}
    \subfigure[Detection Task]{%
    \resizebox{0.48\textwidth}{!}{%
    \begin{tabular}{lrrrrr}
    \toprule
    \textbf{Agent} & \textbf{Accuracy} & \textbf{Time (s)} & \textbf{\# Steps} & \textbf{Input} & \textbf{Output} \\
    \midrule
    \gptfourshell & 69.23\% & 7.08 & 3.85 & 5,492 & 132 \\
    \gptthreeshell & 23.07\% & 11.05 & 13.60 & 1,940.44 & 385.56 \\
    \react & 76.92\% & 39.00 & 11.46 & 15,608.08 & 933.15 \\
    \flash & \cellcolor{Gray}100\% & 78.27 & 6.77 & 12,869.08 & 125.69 \\
    \mksmc & 15.38\% & 1.00 & N/A & N/A & N/A \\
    \bottomrule
    \label{tab:metrics_detection}
    \end{tabular}%
    }}%
    \hfill
    \subfigure[Localization Task]{%
    \resizebox{0.50\textwidth}{!}{%
    \begin{tabular}{lrrrrrr}
    \toprule
    \textbf{Agent} & \textbf{Acc.@3} & \textbf{Acc.@1} & \textbf{Time (s)} & \textbf{\# Steps} & \textbf{Input} & \textbf{Output} \\
    \midrule
    \gptfourshell & 61.54\% & \cellcolor{Gray}61.54\% & 7.04 & 4.23 & 4,588.07 & 133.23 \\
    \gptthreeshell & 30.77\% & 30.77\% & 6.26 & 11.92 & 1,784.23 & 217.08 \\
    \react & \cellcolor{Gray}69.23\% & 53.85\%\textcolor{darkred}{$\downarrow$} & 38.65 & 11.08 & 4,760.77 & 880.92 \\
    \flash & 61.54\% & 46.15\%\textcolor{darkred}{$\downarrow$} & 56.60 & 5.77 & 1,875.08 & 123.31 \\
    \pdiagnose & 15.38\% & 15.38\% & 1.02 & N/A & N/A & N/A \\
    \rmlad & 7.69\% & 7.69\% & 1.98 & N/A & N/A & N/A \\
    \bottomrule
    \label{tab:metrics_localization}
    \end{tabular}%
    }}%
    \vspace{-10pt}
    \subfigure[Root Cause Analysis (RCA) Task]{%
    \resizebox{0.48\textwidth}{!}{%
    \begin{tabular}{lrrrrrr}
    \toprule
    \textbf{Agent} & \textbf{Accuracy} & \textbf{Time (s)} & \textbf{\# Steps} & \textbf{Input} & \textbf{Output} \\
    \midrule
    \gptfourshell & 40.90\% & 8.68 & 4.81 & 4,297.91 & 176.18 \\
    \gptthreeshell & 9.09\% & 10.06 & 14.00 & 1,495.55 & 406.27 \\
    \react & \cellcolor{Gray}45.45\% & 32.16 & 8.00 & 16,276.09 & 757.27 \\
    \flash & 36.36\% & 59.00 & 6.09 & 1,193.90 & 152.45 \\
    \bottomrule
    \label{tab:metrics_rca}
    \end{tabular}%
    }}%
    \hfill
    \subfigure[Mitigation Task]{%
    \resizebox{0.48\textwidth}{!}{%
    \begin{tabular}{lrrrrr}
    \toprule
    \textbf{Agent} & \textbf{Accuracy} & \textbf{Time (s)} & \textbf{\# Steps} & \textbf{Input} & \textbf{Output} \\
    \midrule
    \gptfourshell & 27.27\% & 99.47 & 13.72 & 10,142.55 & 1,060.00 \\
    \gptthreeshell & 0\% & 23.78 & 20.00 & 3,178.33 & 967.71 \\
    \react & 36.36\% & 67.18 & 15.54 & 29,211.90 & 1,464.90 \\
    \flash & \cellcolor{Gray}54.55\% & 216.41 & 16.09 & 8,469.00 & 760.36 \\
    \bottomrule
    \label{tab:metrics_mitigation}
    \end{tabular}%
    }}%
    \vspace{-15pt}
    
\end{table*}

The detection task, being a binary choice question, 
    should be the simplest task
    and the first step an AIOps agent performs.
However, as shown in ~\Cref{tab:metrics_detection},
    only \flash{} answers all the detection problems correctly. 
For localization task,
    agents are allowed to come up with 
    a list of potential faulty services as their answers
    (since there could be multiple faults happenning 
        in the system at the same time). 
To evaluate their accuracy, we consider both the top 1 and top 3 answers.
In ~\Cref{tab:metrics_localization}, 
    \react{} performs best when evaluated using the top 3 answers,
    but its accuracy drops when considering the top 1.
The RCA and mitigation tasks
    prove to be the most challenging for the agents. 
    \gptthreeshell{} fails to recover any failure 
    in its mitigation attempts.

\textit{Problem difficulty differs across task levels.}
    Despite showing promise in addressing realistic operational tasks, 
    none of the agents consistently achieve high problem-solving accuracy 
    across four task categories in \ourbench{} benchmark. 
Even the top-performing agents, such as \flash{}, 
    exhibit limitations, particularly when tackling more complex tasks 
    like mitigation. 
In \Cref{sec:analysis}, we will explore in detail 
    the failure modes and challenges
    contributing to these performance limitations of agents,
    and opportunities for improvement.

\subsection{Influence of the Step Limit}
\label{subsec:increasing_steps}

\begin{figure}[t]
    \centering
    \includegraphics[width=0.95\linewidth]{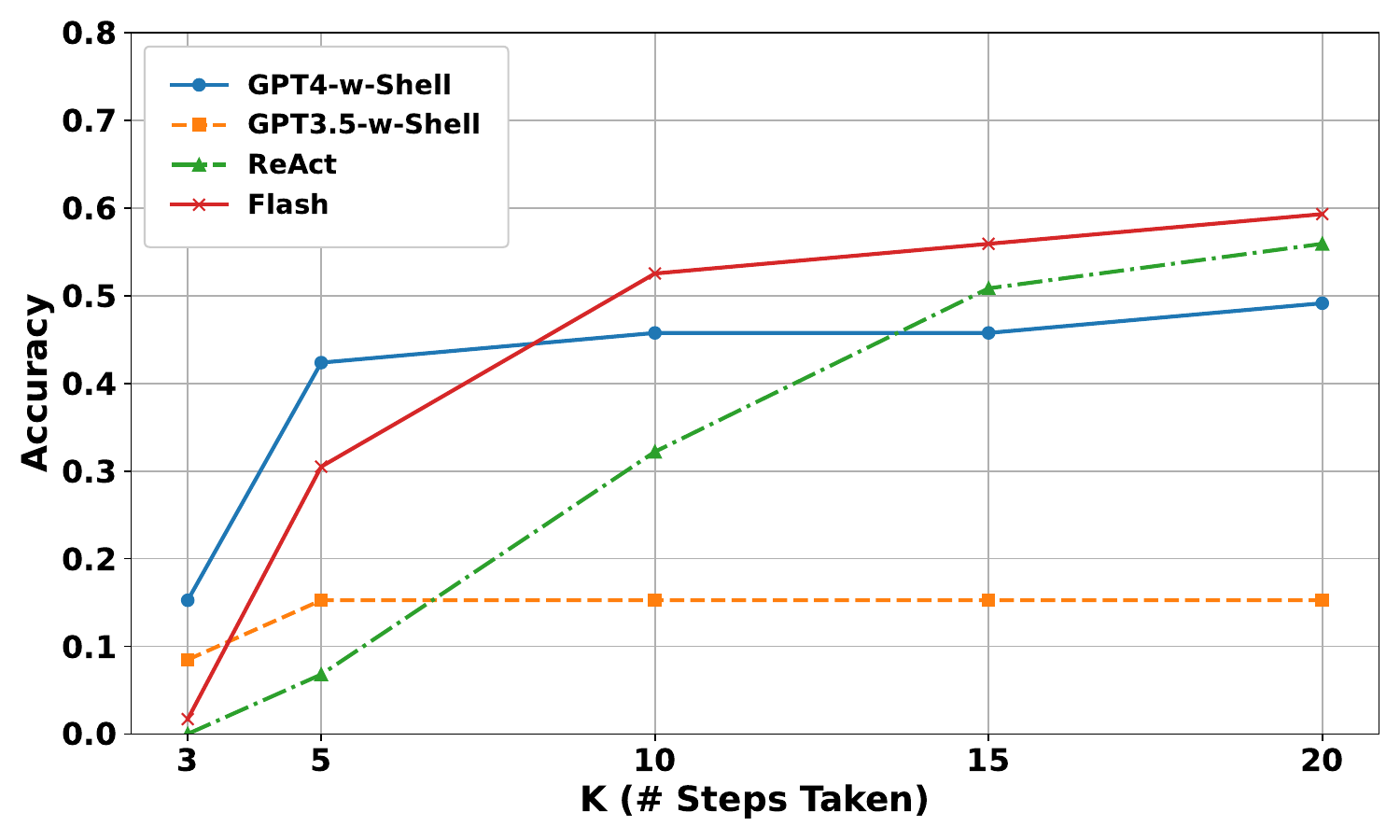}
    \vspace{-10pt}
    \caption{Agent performance vs. number of steps taken.}
    \label{fig:k_steps}
    \vspace{-10pt}
\end{figure}

We examine the impact of the maximum number of 
    allowed steps on the agent's performance, 
    with the results shown in \Cref{fig:k_steps}. 
The step limit significantly affects the performance of certain agents. 
For instance, \react{} and \flash{} show improved accuracy 
    with more steps, with \flash{} reaching the highest accuracy of 59.32\% 
    when the step limit is set to 20. 
However, for \gptthreefiveturbo{}, 
    increasing the step limit beyond 5 does not yield better performance 
    but merely increases the token consumption. 
Notably, the plateauing of accuracy after a certain number of steps indicates that \textit{self-repair with environment feedback can saturate quickly for  AIOps problems}. 
On the contrary, in development tasks (Dev), such as code generation, feedback via various compositional tools such as linters, type checkers, and test cases help agents continuously improve.
This suggests the need for (1) better task decomposition for AIOps problems using planning, (2) improved feedback mechanisms for intermediate steps, and (3) solutions that go beyond environment feedback and self-repair.

\subsection{Agent Behavior: The Good, the Bad and the Gaps}
\label{sec:analysis}

\begin{table}[t!]
\vspace{-0.2in}
\caption{Occurrences of system commands.}
\label{table:other_commands}
\centering
\resizebox{0.47\textwidth}{!}{%
\begin{tabular}{l@{\hskip 4.55pt}c@{\hskip 4.55pt}c@{\hskip 4.55pt}c@{\hskip 4.55pt}c@{\hskip 4.55pt}c@{\hskip 4.55pt}c@{\hskip 4.55pt}c@{\hskip 4.55pt}c@{\hskip 4.55pt}c}
\toprule
\textbf{Agent} & \texttt{find} & \texttt{echo} & \texttt{py} & \texttt{awk} & \texttt{mongo} & \texttt{grep} & \texttt{ls} & \texttt{cat} & \texttt{ip} \\ 
\midrule
\react{} & 0 & 0 & 0 & 3 & 0 & 1 & 26 & 30 & 0 \\ 
\flash{} & 0 & 3 & 0 & 0 & 0 & 0 & 8 & 10 & 0 \\ 
\bottomrule
\end{tabular}
}
\end{table}

\begin{figure}[t]
\vspace{-0.15in}
    \centering
    \includegraphics[width=1\linewidth]{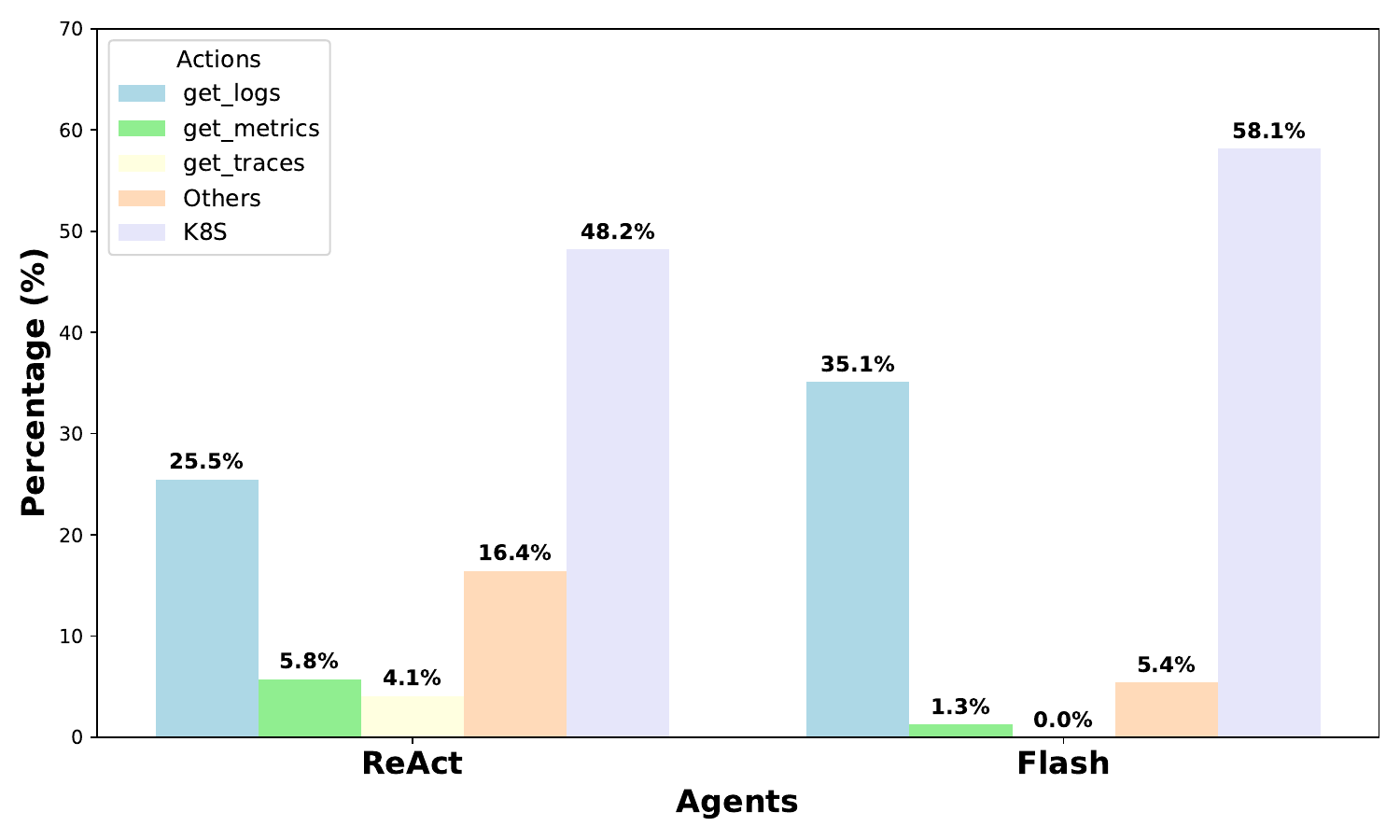}
    \vspace{-20pt}
    \caption{Total percentage of actions taken by different agents.}
    \label{fig:actions_distribution}
    \vspace{-10pt}
\end{figure}

We now delve into the behaviors of the agents 
    and analyze the good, the challenges, 
    and opportunities for improvement.
In \Cref{table:agent_performance}, 
    we see that all agents perform 
    better than the traditional non-LLM AIOps methods
    in terms of the problems for detection and localization tasks.
\Cref{fig:actions_distribution}, 
    shows the
    telemetry API usage patterns among agents.
The \getlogs{} API
    is the most frequently used API across all agents,
    then the \getmetrics{}, and the \gettraces{} APIs.
However, agents also diverge in their patterns of API usage.
For example, 
\flash{} does not use the \gettraces{} API at all.
We present the occurrences of other system commands 
    for each agent in \Cref{table:other_commands}.
We next discuss the underlying reasons and patterns contributing to the agents' poor performance.

\subsubsection{Wasting steps on unnecessary actions}
\label{subsubsec:unnecessary_actions}
We observe that agents often waste steps on unnecessary actions,
    such as repeatedly calling the same API, 
    generating non-existent APIs,
    or spending excessive steps in multi-agent communication.
Specifically, the \gptthreeshell{} agent often generates incorrect API commands in loops,
    leading to repeated errors in execution.
For instance, setting \texttt{\small speaker\_selection\_method}
    as \texttt{\small round\_robin} allows every agent to speak in every step,
    but this often prevents decisive, efficient decisions,
    as agents repeatedly resort to telemetry APIs for more information.
Even with the \texttt{\small speaker\_selection\_method}
    set to \texttt{\small auto}, where the next speaker is automatically chosen, 
    a selected agent always
     speaks ten times in a step without communication
    (with a maximum of ten communication rounds per step).

\begin{figure}[t]
    \centering
    \vspace{-5pt}
    \subfigure[Successful cases.]{%
        \includegraphics[width=0.40\columnwidth]{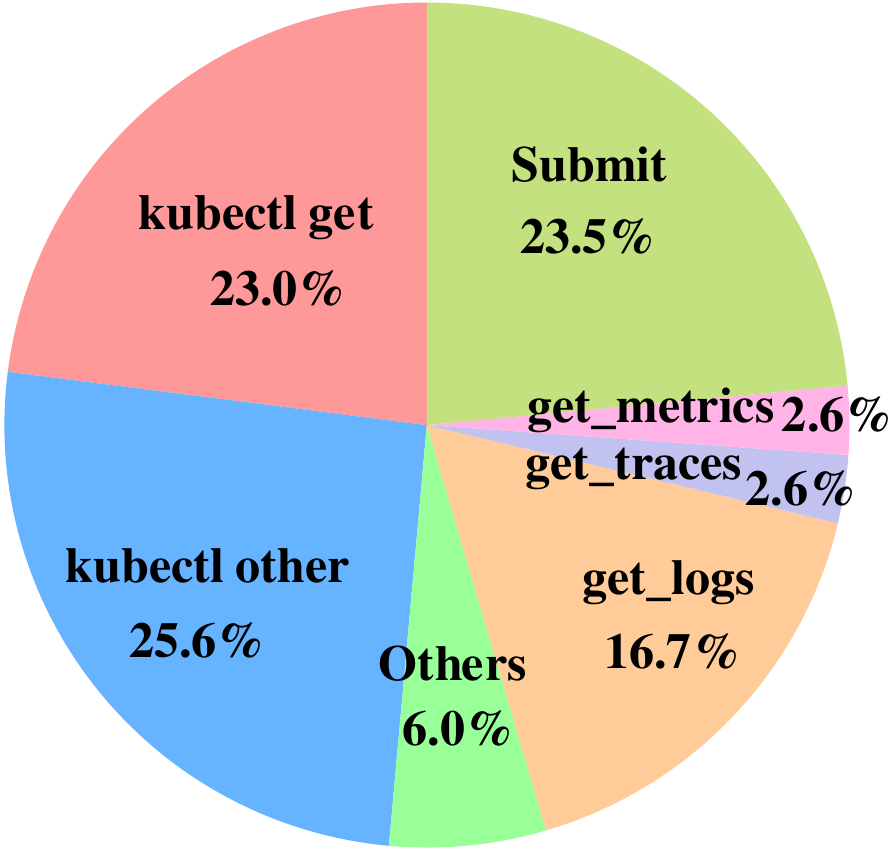}%
        \label{fig:success_action}
    }
    \subfigure[Failure cases.]{%
        \includegraphics[width=0.40\columnwidth]{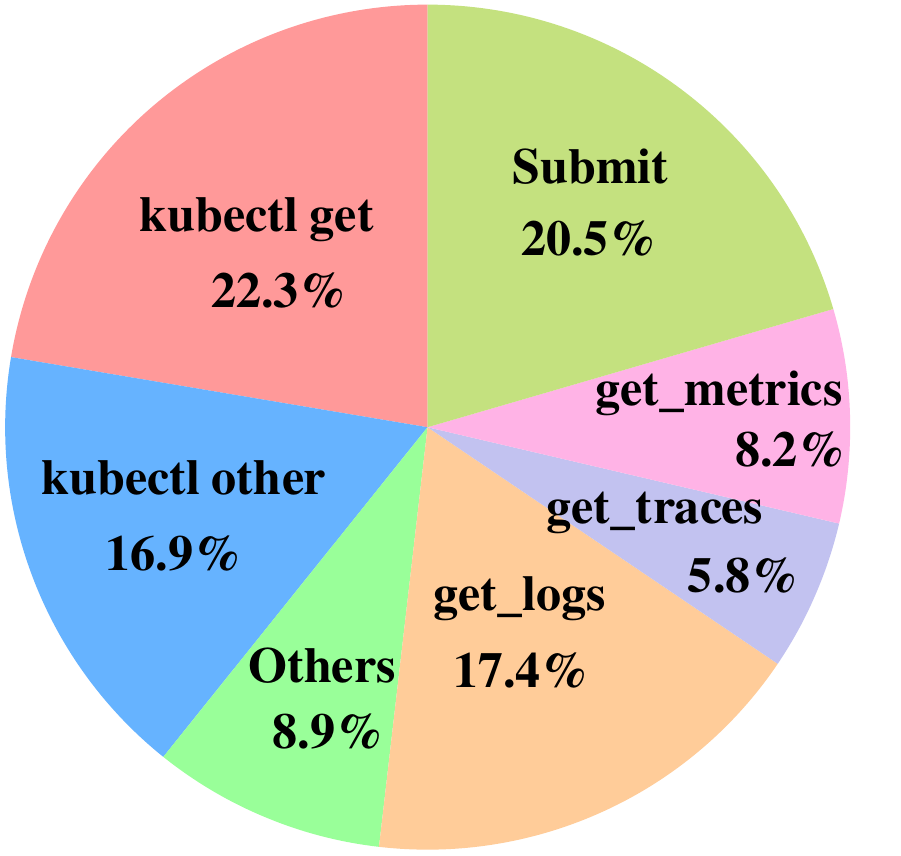}%
        \label{fig:failure_action}
    }
    \vspace{-10pt}
    \caption{Action distribution by success and failure cases.}
    \label{fig:action_cases}
    \vspace{-0.15in}
\end{figure}

\subsubsection{Overloaded information when consuming data}
\label{subsubsec:telemetry}
To dig deeper into the agent failure modes, 
    we analyze the correlation between the agents' actions and the success or failure of problem-solving, 
    as well as the distribution of actions across steps.
In \Cref{fig:action_cases}, 
    we present the distribution of actions 
    for both successful and failed cases.
Agents tend to 
    use \getmetrics{} and \gettraces{} APIs sparingly in successfully resolved problems, 
    typically only when necessary.
This is understandable, as the metrics data, 
    e.g., CPU and memory usage have numerous values,
    which are hard to directly interpret,
    and trace data are descriptive records 
    of the system's dependencies,
    which are more comprehensible when visualized.
However, agents may subsequently consume
    these data with a \code{cat}
    command directly, which can overwhelm
    the model's input context window
    and cause distraction and more tokens to be consumed.
Consequently, using these telemetry APIs 
    without careful consideration 
    or analysis can add more noise
    into the agents' reasoning,
    possibly leading to token exhaustion.
We expect more refined telemetry data processing 
    and filtering mechanisms to be implemented in the agents
    to avoid this issue in the future.

\subsubsection{Invalid API usage}
\label{subsubsec:invalid_api}
We notice that agents can struggle with improper formatting of API calls.
For instance, \gptthreeshell{} consistently generates 
    incorrect command formats (though the API name is correct),
    such as malformed parameters, 
    and repeat the error in subsequent steps.
In many cases, \gptthreeshell{} 
    repeatedly responds with: 
    ``I apologize for the error. Here is the API call again:''
    followed by the same/similar mistake. 
This issue is observed as many as 14 times 
    within a 20-step problem-solving case.

Similar issues with invalid API usage are also observed in other agents.
For instance, the \react{} agent occasionally generates incorrect API commands,
    but typically recovers by reasoning 
    through the errors and self-correcting its commands. 
In the following example,
    \react{} uses an incorrect parameter for the 
    \code{get_logs} API, but subsequently corrects it by 
    checking the existing services in its next step.
\begin{tcolorbox}[
    fontupper=\scriptsize,
    fonttitle=\scriptsize,
    sharp corners,               
    boxrule=0.5pt,
    boxsep=1.5pt,
    left=2pt,             
    right=2pt,        
    top=2pt,
    bottom=2pt 
    ]
\textbf{\reactname{}}: \texttt{get\_logs("test-social-network", \textcolor{darkpastelred}{"Social Network"})} \\
\textbf{\ourbench{}}: Error: Your service/namespace does not exist. \\
\textbf{\reactname{}}: \\
\hspace*{1em}\textit{Thought:} I should check the services deployed within the test-social-network namespace to ensure I use the correct service name. \\
\hspace*{1em}\textit{Action:} I will list the services in the test-social-network namespace to confirm the correct name of the service. \\
\hspace*{1em}\texttt{exec\_shell("kubectl get services -n test-social-network")}
\end{tcolorbox}

\subsubsection{False positive detection issues}
\label{subsubsec:false_positive}
To further evaluate the agents' performance, 
    we set up two detection problems for the two microservice applications 
    where no faults exist,
    referred to as no operation (Fault 10 -- Noop in \Cref{tbl:faults}) problems.
Only \gptfourshell{}  
    correctly identify these cases 
    as normal system execution,
    while the others report false positives,
    misinterpreting normal activities 
    (e.g., standard workload generation) as faults.

\section{Discussion}
\label{sec:discussions}

\ourbench{} helps engineers to easily create customized incident scenarios for evaluating agents. By providing Agent Cloud Interfaces (ACIs) as guard-rails, \ourbench{} ensures that agents are tested within a controlled environment, allowing users to focus on designing scenarios that accurately represent incidents in their systems and defining the specific problems their agents should solve.

\ourbench{} is also adaptable to other fault types. For example, an anomaly detection workload scenario can be introduced for detection tasks. Further, users can create problems where agents are required to label the workload or telemetry data to identify anomalies.

When implementing problem evaluators, fine-grained evaluation oracles, or \ourbench{}'s optional LLM-as-Judge, may be necessary. For instance, in the binary-choice detection task, agents may answer correctly but provide incorrect interpretations or reasoning. In one case, an agent claimed to detect an abnormal system behavior, but its explanation referenced a workload that was, in fact, normal and unrelated to the injected fault. Leveraging \ourbench{}'s LLM-as-Judges can help address this issue by comparing the LLM reasoning chains with the problem description (including the fault, workload, and environment setup).

\section{Related Work}
\label{sec:related}
\para{AgentOps.} Recent advancements in cloud management have increasingly incorporated LLMs to enhance operational tasks.
Approaches such as fine-tuned GPT models \cite{ahmed2023recommending}, RCACopilot \cite{chen2024automatic}, RCAgent \cite{wang2023rcagent}, MonitorAssistant \cite{yu2024monitorassistant}, and Xpert \cite{jiang2024xpert} illustrate the effectiveness of LLMs in monitoring and analyzing complex system behaviors.
However, beyond the lack of publicly available implementations and associated private datasets, 
there is a notable gap: the absence of a unified benchmark capable of providing realistic evaluation scenarios to assess agents' performance across operational tasks.

\para{AIOps benchmarks.} 
Existing AIOps benchmarks primarily rely on 
    static or text-based datasets,
    such as system metrics~\cite{adbench,anomalybench}, 
    typically time series data, 
    or fixed question-answer format~\cite{opseval}.
These benchmarks, together with the general Language Model benchmarks~\cite{hendryckstest2021,hendrycks2021ethics,liang2023holisticevaluationlanguagemodels,lee2023holisticevaluationtexttoimagemodels,srivastava2023beyond,huang2023ceval},
    do not simulate the dynamic
    and complex cloud environments, 
    not to mention allowing agents 
    to interact with them to solve operational tasks.

\section{Conclusion}
\label{sec:conclusion}
In this paper, we unravel the requirements and challenges for a comprehensive framework that supports the design, development, and evaluation of autonomous AIOps agents.
We develop a framework, \ourbench{}, which combines a fault injector, workload generator, cloud-agent orchestrator, and telemetry observer to simulate cloud incidents and provide an agent-cloud interface for orchestrating and evaluating AIOps agents.
We leverage \ourbench{} to construct a benchmark suite with 48 problems and evaluate four agents to demonstrate the application of \ourbench{} in evaluating LLM-based agents across different types of AIOps tasks.

\bibliographystyle{ACM-Reference-Format}
\bibliography{references}

\end{document}